\documentclass[letterpaper]{article} % DO NOT CHANGE THIS
\usepackage{aaai2026}
\usepackage{times}  % DO NOT CHANGE THIS
\usepackage{helvet}  % DO NOT CHANGE THIS
\usepackage{courier}  % DO NOT CHANGE THIS
\usepackage[hyphens]{url}  % DO NOT CHANGE THIS
\usepackage{graphicx} % DO NOT CHANGE THIS
\usepackage{booktabs} % cm added
\usepackage{multirow} % cm added
\usepackage{natbib}  % DO NOT CHANGE THIS AND DO NOT ADD ANY OPTIONS TO IT
\usepackage{caption} % DO NOT CHANGE THIS AND DO NOT ADD ANY OPTIONS TO IT
\usepackage{algorithm}
\usepackage{algpseudocode}
\usepackage{amsmath}
\usepackage{tcolorbox}
\usepackage{amsfonts} % for \mathbb
\usepackage{makecell}
\usepackage{pifont}
\usepackage{bm}

\makeatletter
\def\copyright@on{} % turn off copyright
\makeatother

\usepackage{newfloat}
\usepackage{listings}
\DeclareCaptionStyle{ruled}{labelfont=normalfont,labelsep=colon,strut=off} % DO NOT CHANGE THIS
\floatstyle{ruled}
\newfloat{listing}{tb}{lst}{}
\floatname{listing}{Listing}
\title{GraphCoT-VLA: A 3D Spatial-Aware Reasoning Vision-Language-Action Model for Robotic Manipulation with Ambiguous Instructions}
\author{
    Helong Huang\textsuperscript{\rm 1}\equalcontrib,
    Min Cen\textsuperscript{\rm 2}\equalcontrib\thanks{Corresponding author.
    },
    Kai Tan\textsuperscript{\rm 1},
    Xingyue Quan\textsuperscript{\rm 1},
    Guowei Huang\textsuperscript{\rm 1},
    Hong Zhang\textsuperscript{\rm 3}
}
\affiliations{
    \textsuperscript{\rm 1}Noah’s Ark Lab, Huawei\\ 
    \textsuperscript{\rm 2}School of Artificial Intelligence and Data Science, University of Science and Technology of China\\
    \textsuperscript{\rm 3}School of Management, University of Science and Technology of China
}

\usepackage{bibentry}
\begin{document}

\maketitle

\begin{abstract}

Vision-language-action models have emerged as a crucial paradigm in robotic manipulation. However, existing VLA models exhibit notable limitations in handling ambiguous language instructions and unknown environmental states. Furthermore, their perception is largely constrained to static two-dimensional observations, lacking the capability to model three-dimensional interactions between the robot and its environment. To address these challenges, this paper proposes GraphCoT-VLA, an efficient end-to-end model. To enhance the model's ability to interpret ambiguous instructions and improve task planning, we design a structured Chain-of-Thought reasoning module that integrates high-level task understanding and planning, failed task feedback, and low-level imaginative reasoning about future object positions and robot actions. Additionally, we construct a real-time updatable 3D Pose-Object graph, which captures the spatial configuration of robot joints and the topological relationships between objects in 3D space, enabling the model to better understand and manipulate their interactions. We further integrates a dropout hybrid reasoning strategy to achieve efficient control outputs. Experimental results across multiple real-world robotic tasks demonstrate that GraphCoT-VLA significantly outperforms existing methods in terms of task success rate and response speed, exhibiting strong generalization and robustness in open environments and under uncertain instructions.

% Our approach introduces a novel structured Chain-of-Thought reasoning module, which integrates high-level task understanding and planning, failed task feedback, and low-level imaginative reasoning about future object position and robot actions. This design enhances the model's ability to interpret fuzzy instructions and improve task planning.
% Additionally, we construct a real-time updatable 3D Pose-Object graph that explicitly models the spatial poses of robot joints alongside the spatial positions and geometric relationships of objects in the environment. This enables the model to better understand and manipulate the three-dimensional interaction process between the robot and objects. 

% 视觉-语言-动作模型（Vision-Language-Action models, VLAs）逐渐成为机器人领域的重要范式。然而，现有 VLA 模型在处理模糊语言指令和未知环境状态时仍存在显著不足。此外，其感知能力大多局限于静态二维观测，缺乏对机器人与环境间三维交互关系的建模能力。为此，本文提出 GraphCoT-VLA，一种高效的端到端模型。模型提出了全新的结构化的 Chain-of-Thought（CoT）推理模块，包括高层次的任务理解和规划，失败任务反馈，以及低层次的对未来物体状态和机器人动作的想象推理，从而提升对模糊指令的理解与任务规划能力；其次构建了可实时更新的3D空间下的 Pose-Object 图增强模型理解，通过融合机器人关节位姿、环境中物体的空间位置与几何关系，显式建模机器人与物体之间的三维交互过程，增强模型的空间理解与操控能力。我们还提出 Dropout 混合推理策略，实现低延迟、高频率的控制输出。在多项真实机器人任务中，GraphCoT-VLA 在任务成功率与响应速度方面显著优于现有方法，展现出在开放环境和不确定指令下的强泛化能力与鲁棒性。

\end{abstract}


\begin{figure}[t]
  \centering
  \includegraphics[width=\linewidth]{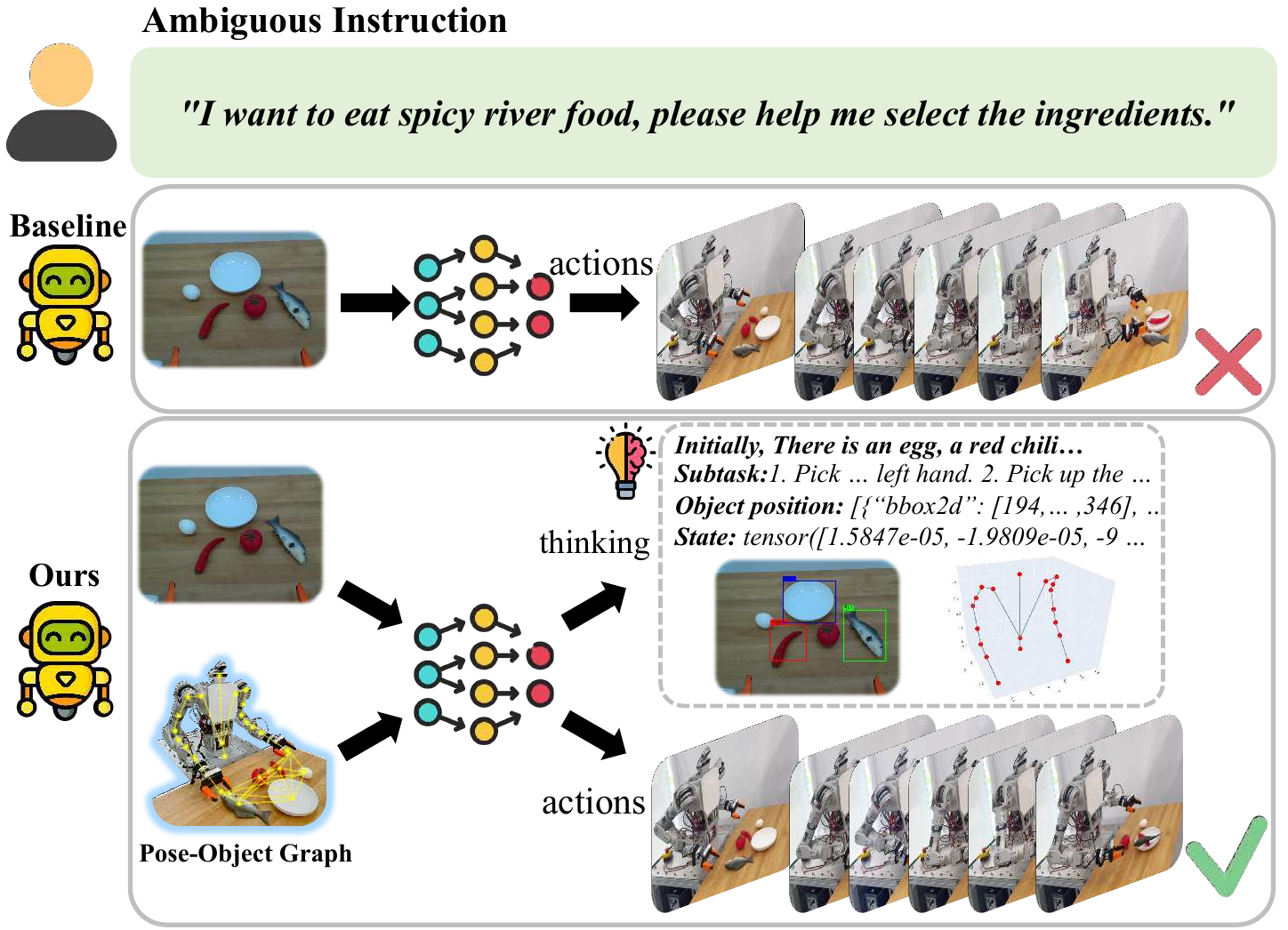}
  \caption{Comparison between our method and baseline.}
  \label{fig:introduction}
\end{figure}

\begin{figure*}[t]
    \centering
    \includegraphics[width=\textwidth]{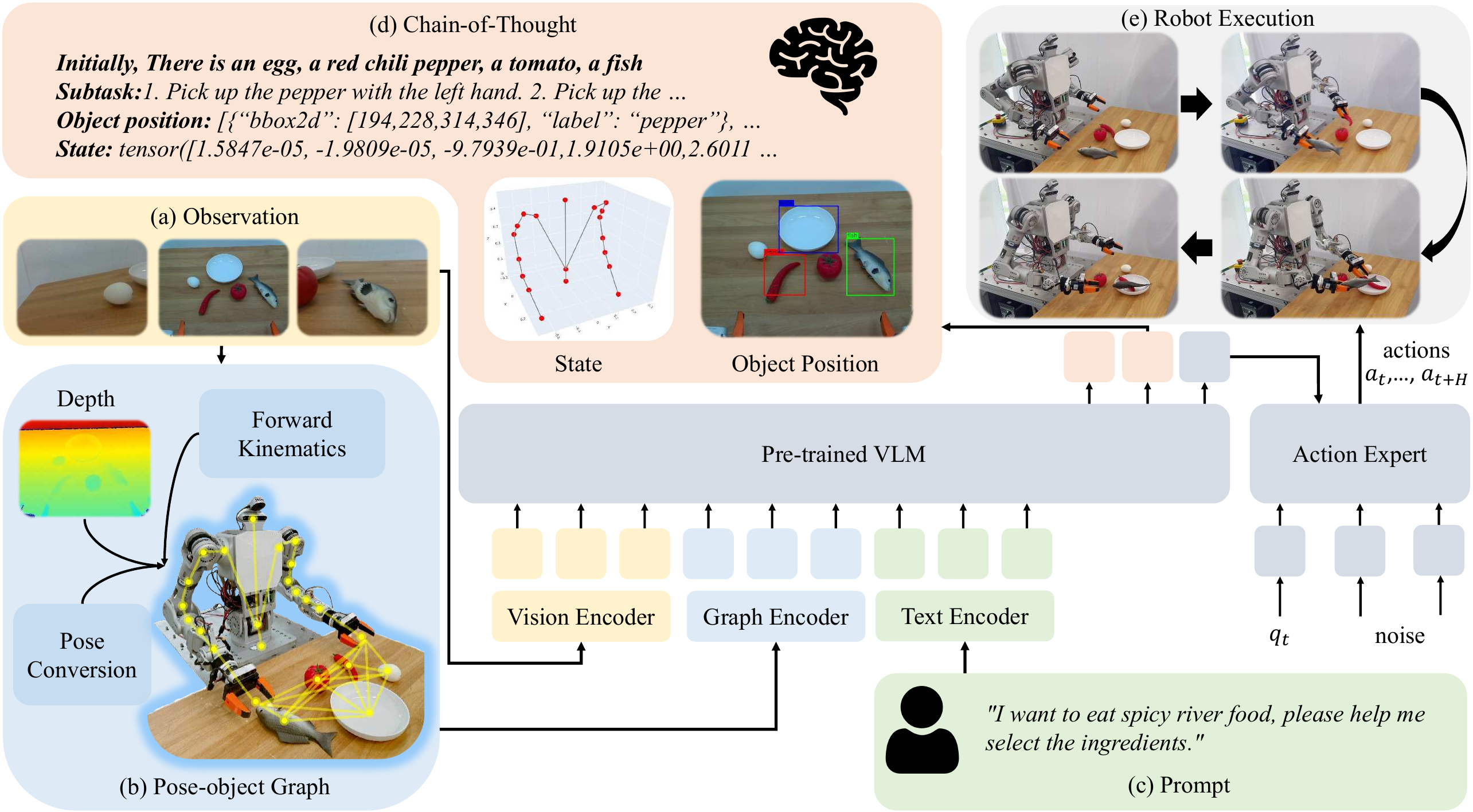}
    \caption{\textbf{Overview of GraphCoT-VLA.} GraphCoT-VLA takes multi-view visual observations, an ambiguous language instruction, and a 3D Pose-Object Graph as input. The graph is constructed using robot kinematics and depth data to represent 3D spatial relationships in the scene. A structured Chain-of-Thought reasoning process enables scene understanding, subtask decomposition, and imagination of object states and positions, which are autoregressively generated to guide the action expert.}
    % (a) Observation: We use cameras from three different perspectives to collect visual observations. 
    % (b) Pose-Object Graph: We use forward kinematics to compute the 3D poses of robot links, detect object positions, estimate distances from depth images, and transform all poses into the robot base frame using extrinsic calibration. 
    % (c) Prompt: A ambiguous language instruction is provided as the task prompt. 
    % (d) Chain-of-Thought: We design the Chain-of-Thought reasoning process to guide the VLM, which includes: (1) high-level task and scene understanding, (2) subtask decomposition for planning, (3) prediction of target object localization, and (4) prediction of target joint angles. (e) Robot Execution: Execution on real robot.} 
    \label{fig:overview}
\end{figure*}

\section{Introduction}

% VLAs introduction
To achieve natural and efficient human-robot interaction in the real world, end-to-end vision-language-action (VLA) models are increasingly becoming a central paradigm in robotics~\cite{black2024pi_0, liu2024rdt, wen2024diffusion, wen2025dexvla, zhang2023crossformer, team2024octo}. VLA models interpret and execute human instructions expressed in natural language within a closed-loop system that integrates perception, understanding, and action~\cite{din2025vision}. Many existing approaches build upon pre-trained Vision-Language Models (VLMs) to form Vision-Language Action (VLA) frameworks, leveraging VLMs' strong visual grounding and language understanding capabilities to jointly encode multimodal inputs and directly predict control actions for robotic manipulation \cite{black2024pi_0,kim2024openvla}. 
% For instance, $\pi_0$~\cite{black2024pi_0} employs a Vision Transformer (ViT)~\cite{dosovitskiy2020image} to encode visual inputs, which are then combined with language instructions to generate actions through Flow Matching.

% However, in real-world scenarios, users often lack full awareness of the current environment state and struggle to provide detailed or accurate instructions. a user might say \textit{“I want to eat spicy river food”}, without knowing whether the necessary ingredients are available in the refrigerator. Such vague commands place greater demands on intelligent agents. The model must be capable of inferring concrete, executable low-level actions based on multimodal perceptual inputs, while also providing feedback or corrections based on environmental constraints. Most existing VLA models, however, rely on clear and structured language instructions to perform specific tasks~\cite{black2024pi_0, kim2024openvla, liu2024rdt, team2024octo}. When faced with ambiguous commands or instructions that exceed the capabilities of the environment, these models often either fail to execute or generate unrealistic behaviors and explanations. This reflects a fundamental limitation: they lack the ability to align their reasoning and actions with the physical environment.

Most existing VLA models rely on clear and structured language instructions to perform specific tasks \cite{black2024pi_0, kim2024openvla, liu2024rdt}. However, in real-world scenarios, users often lack full awareness of the environment and may issue vague commands like ``I want to eat spicy river food.'' without knowing whether the necessary ingredients are available. When encountering such ambiguous or infeasible instructions, these models often fail or produce hallucinated actions, revealing a fundamental flaw in existing approaches: their inability to contextualize multimodal perceptual inputs and adapt action planning accordingly, which prevents reasoning and decision-making from being grounded in the real physical environment. Some works have introduced Chain-of-Thought (CoT) \cite{wei2022chain} with VLA models to improve reasoning ability. ECoT \cite{zawalski2024robotic} decomposes tasks into multi-step subtasks for planning, but struggles with ambiguous commands and unmet user expectations due to limited situational awareness. CoT-VLA \cite{zhao2025cot} imagines observations of subtasks, yet lacks the ability to plan or give feedback based on a full understanding of the scene.

In addition, existing VLA models remain limited in observation modalities, typically relying only on RGB images from multiple viewpoints~\cite{black2024pi_0, zhao2025cot}. This restricted perception scope hinders model's ability to develop a deep understanding of the environment. To improve this, recent studies have explored the integration of additional modalities, such as positional encodings~\cite{qu2025spatialvla} and depth information~\cite{wang2025roboflamingo} to enhance spatial perception . Although these methods expand the input space, they primarily rely on static 2D views of individual objects and lack the capacity to model complex 3D interactions between the robot and its environment. 

% To address the two challenges mentioned above, we propose GraphCoT-VLA, an efficient end-to-end model, as illustrated in Fig.~\ref{fig:introduction}, the model introduces a novel CoT architecture that dynamically analyzes current observations, interprets ambiguous instructions, and generates failure feedback. It also possesses the ability to predict future object states and robot actions, thereby enhancing reasoning and decision-making under vague commands and uncertain environments. n particular, we introduce a real-time Pose-Object Interaction Graph that explicitly encodes the evolving 3D spatial relationships between the robot and surrounding objects, facilitating fine-grained perception and interaction modeling in continuous space. This strengthens the model's perception and execution capabilities for complex manipulation tasks. To balance performance and speed, GraphCoT-VLA incorporates a Dropout-based joint training mechanism and a hybrid reasoning strategy, enabling high-frequency, low-latency real-time control. The main contributions of this paper are as follows:

To address the above challenges, we propose GraphCoT-VLA, an efficient end-to-end model for robotic manipulation. As shown in Fig.~\ref{fig:introduction}. the core of our model is a novel CoT architecture that performs dynamic observation analysis, interprets ambiguous instructions, generates failure feedback, and predicts future object states and robot actions, enabling robust reasoning and temporal forecasting in uncertain environments. We further introduce a real-time Pose-Object Graph to explicitly model 3D spatial relations between the robot and objects, enhancing perception and understanding of 3D space. To ensure real-time performance, GraphCoT-VLA integrates a dropout-based joint training strategy~\cite{chen2025training} and a hybrid CoT reasoning mechanism, balancing fast inference with deep reasoning. Our key contributions are:
% \begin{itemize}
%     \item A novel structured CoT reasoning module that enhances the model’s task understanding and feedback capabilities in the presence of ambiguous instructions and open-world scenarios.
%     \item We design a real-time updatable Pose-Object Graph to improve the model’s understanding of the three-dimensional interaction between the robot body and objects.
%     \item We introduce a Dropout-based hybrid reasoning strategy to enable efficient and stable generation of both reasoning text and actions, supporting feedback generation and real-time robot control.
%     \item We validate the proposed model on real robotic platforms, demonstrating that it significantly outperforms existing methods across multiple tasks, with notable improvements in both accuracy and response speed.
% \end{itemize}
\begin{itemize}
    % \item GraphCoT-VLA is proposed, an novel end-to-end model for robotic manipulation under ambiguous instructions and open-world scenarios.
    \item A novel end-to-end model, GraphCoT-VLA, is proposed for robotic manipulation under ambiguous instructions and open-world conditions.
    % \item A novel structured CoT module that supports scene understanding, feedback generation, and future prediction.
    \item A structured CoT module is proposed to support scene understanding, feedback, and future imagination.
    % \item A real-time Pose-Object Graph is designed to model dynamic 3D interactions between the robot and surrounding objects.
    \item A real-time Pose-Object Graph is designed to model 3D interactions between the robot and surrounding objects.
    \item A hybrid CoT reasoning strategy using dropout-based training enable both fast inference and iterative refinement for real-time control.
    \item GraphCoT-VLA is validated on real robots, demonstrating superior performance in success rate, action fluency, temporal modeling, and task generalization.
\end{itemize}

\section{Related Work}

\paragraph{Vision-Language-Action Models.} VLA models are increasingly emerging as a unified framework that enables robots to perform complex manipulation tasks~\cite{black2024pi_0, kim2024openvla, wen2025dexvla, zhang2023crossformer, team2024octo}. Some approaches adopt Transformer-based~\cite{vaswani2017attention} architectures to generate control commands~\cite{brohan2022rt, wu2023unleashing, cheang2024gr}, while others leverage the enhanced capabilities of VLMs~\cite{zitkovich2023rt, wen2025tinyvla}. For instance, $\pi_0$~\cite{black2024pi_0} fine-tunes Gemma to generate continuous actions via flow matching~\cite{lipman2022flow}. Diffusion-based models have also emerged as promising solutions: RDT-1B~\cite{liu2024rdt} employs a diffusion Transformer to generate dual-arm control vectors. Other methods aim to improve cross-signal generalization. For example, DexVLA~\cite{wen2025dexvla} and Crossformer~\cite{zhang2023crossformer} are designed to handle diverse robotic platforms. In contrast to prior methods that are limited to executing explicit commands, our goal is to enhance task comprehension and feedback responsiveness in open-world scenarios with ambiguous instructions.

\begin{figure}[t]
  \centering
  \includegraphics[width=\linewidth]{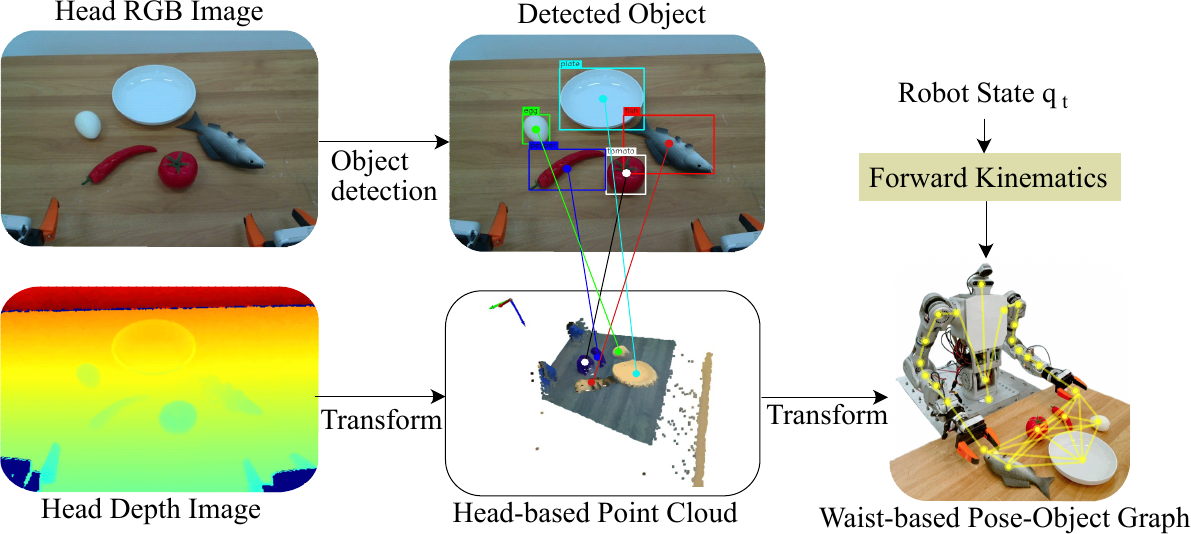}
  \caption{Pose-Object Graph generation.}
  \label{fig:pose_graph_gen}
\end{figure}

\paragraph{Enhance the Reasoning and Comprehension Abilities of VLA.} Although some VLA models leverage VLMs, they still lack explicit temporal planning and reasoning capabilities. Recent approaches enhance reasoning by introducing CoT mechanisms. For example, CoT-VLA~\cite{zhao2025cot} employs causal attention to generate subgoal images, ECoT~\cite{zawalski2024robotic} incorporates intermediate reasoning steps to understand task structure. For perception, some works integrate richer information to improve understanding. SpatialVLA~\cite{qu2025spatialvla} incorporates positional encodings to capture key spatial points, while RoboFlamingo-Plus~\cite{wang2025roboflamingo} fuses RGB and depth images to enhance perception. However, existing methods still lack dynamic planning and feedback mechanisms when faced with ambiguous instructions. Most are limited to static 2D observations of single objects, making it difficult to model 3D interactions between the robot and its environment. Therefore, this study proposes a novel structured CoT and 3D spatial graph framework to improve the model's understanding and reasoning of ambiguous scenes.

\section{Methods}

% helong
% 这一节将介绍我们整体的方法，我们的将会介绍Pose-Object graph的相关信息, 紧接着介绍CoT的设计方法，在此基础上介绍我们方法的整体架构，以及我们加上CoT的训练策略。
This section presents our overall approach, we first introduce the Pose-Object Graph, then describe the design of CoT, followed by the overall architecture and co-training strategy.

\subsection{Pose-Object Graph}

% helong 
\paragraph{Real-time Construction.} 
To model 3D spatial relationships between the robot and scene objects, we construct a real-time Pose-Object Graph $G_i$ at each time step $t_i$, using synchronized multimodal data $\mathcal{D} = \{ (t_i, I^{\mathrm{rgb}}_i, I^{\mathrm{depth}}_i, \mathbf{q}_i) \}_{i=1}^N$, where $t_i$ is the timestamp, $I^{\mathrm{rgb}}_i$ and $I^{\mathrm{depth}}_i$ are the RGB and depth images, and $\mathbf{q}_i$ is the $J$-DOF joint configuration. We apply YOLO-World~\cite{cheng2024yolo} to detect $n$ object bounding boxes from $I^{\mathrm{rgb}}_i$, extract their 2D centers and project them to 3D points in the head camera frame using depth values and intrinsics $K$, then transform them into the robot base frame using extrinsics $T$. The robot’s end-effector positions are computed via forward kinematics from $\mathbf{q}_i$. All object and end-effector nodes are fully connected to form $G_i$, which is provided to the VLA model for enhanced 3D scene understanding. Fig.~\ref{fig:pose_graph_gen} illustrates the process and details are given in Algorithm~\ref{alg:pose_object_graph}.

\begin{algorithm}[t]
\caption{Real-time Pose-Object Graph Construction.}
\label{alg:pose_object_graph}
\textbf{Input:} Demonstration $\mathcal{D} = \{ (t_i, I_i^{\mathrm{rgb}},
I_i^{\mathrm{depth}}, \mathbf{q}_i) \}_{i=1}^N$, intrinsics matrix $K$,  extrinsic matrix $T$ \\
\textbf{Output:} The corresponding generated graphs $\{G_i\}_{i=1}^N$
\begin{algorithmic}
\For{$i = 1$ to $N$}  \Comment{Same logic used during inference}
    \State $\mathcal{B}_i \leftarrow \mathrm{DetectObj}(I_i^{\mathrm{rgb}})$
    \State $\{(x_j, y_j)\}_{j=1}^n \leftarrow \mathcal{F}_\mathrm{center}(\mathcal{B}_i)$
    \State Initialize $\mathcal{V}_\mathrm{obj} \leftarrow \emptyset$
    \For{$j = 1$ to $n$}
        \State $D_j \leftarrow \mathcal{F}_\mathrm{depth}(I_i^{\mathrm{depth}}, (x_j, y_j))$
        \State $P_j^{\mathrm{head}} \leftarrow D_j \cdot K^{-1} (x_j, y_j, 1)^\top$        
        \State $P_j^{\mathrm{base}} \leftarrow T \cdot P_j^{\mathrm{head}}$  \Comment{Convert to robot base frame}
        \State $\mathcal{V}_\mathrm{obj} \leftarrow \mathcal{V}_\mathrm{obj} \cup \{ P_j^{\mathrm{base}} \}$
    \EndFor
    \State $\mathcal{V}_\mathrm{ee} \leftarrow \mathrm{FK}(\mathbf{q}_i)$
    \State $E \leftarrow \{(v_\mathrm{obj}, v_\mathrm{ee}) \mid v_\mathrm{obj} \in \mathcal{V}_\mathrm{obj}, v_\mathrm{ee} \in \mathcal{V}_\mathrm{ee} \}$
    \State $G_i \leftarrow (\mathcal{V}_\mathrm{obj} \cup \mathcal{V}_\mathrm{ee}, E)$
\EndFor
\State \Return $\{G_i\}_{i=1}^N$
\end{algorithmic}
\end{algorithm}

\paragraph{Graph Encoder.} We encode the Pose-Object Graph $G_t$ using a two-layer graph neural network (GNN)~\cite{scarselli2008graph}, where each layer sequentially applies layer normalization, graph convolution, and ReLU activation. Let $\mathbf{H}^{(0)} \in \mathbb{R}^{(m+n) \times d}$ denote the initial embeddings of Pose-Object Graph, represented as 3D positions in a $d$-dimensional space. We update feature by:
\begin{equation}
\mathbf{H}^{(l+1)} = \sigma \big( \mathcal{G}(\mathcal{N}(\mathbf{H}^{(l)}), \mathbf{A}) \big),
\end{equation}
where $\mathcal{N}$, $\mathcal{G}$, and $\sigma$ denote layer normalization, graph convolution, and ReLU activation, and $\mathbf{A}$ is the adjacency matrix of $G_t$. The final node features are fed into the VLM for downstream reasoning and action prediction.

\subsection{CoT Reasoning via Understanding, Feedback and Imagination}
% To enhance the model's comprehension and reasoning capabilities under ambiguous instructions and previously unseen visual observations, and to enable suggestion-based feedback within the VLA framework, we propose a novel Structural Chain-of-Thought (CoT) mechanism (Figure \ref{fig:overview}). This reasoning chain integrates real-time scene understanding, task instruction interpretation, error detection and suggestion generation, as well as imagination of object positions and future states.

To improve reasoning under ambiguous instructions and novel visuals, and enable suggestion-based feedback in the VLA framework, we propose a structural CoT mechanism (Fig.~\ref{fig:overview}) that integrates scene understanding, instruction interpretation, feedback generation, and future imagination.

Specifically, we first use Qwen2.5-VL~\cite{bai2025qwen2} to perform initial scene understanding from the head image by identifying objects in the environment, prompt example:
\begin{tcolorbox}[colback=gray!5!white, colframe=black, boxrule=0.5pt, arc=2pt]
The desk contains the necessary ingredients and a plate. Please observe the image and describe the items on the desk with very simple sentences.
\end{tcolorbox}

Next, guided by task requirements, we perform feasibility analysis and decompose ambiguous instructions. If the scene does not satisfy the instruction, we query Qwen2.5-VL with a targeted prompt to generate feedback and suggestions, prompt example: 
\begin{tcolorbox}[colback=gray!5!white, colframe=black, boxrule=0.5pt, arc=2pt]
I want to eat spicy river seafood. Please recommend the ingredients that are still missing from the table and needed with very simple sentences.
\end{tcolorbox}

% Specifically, we first employ Qwen2.5-VL-7B to perform an initial scene understanding of the head image, identifying simple objects in the environment.
% \begin{tcolorbox}[colback=gray!5!white, colframe=black, boxrule=0.5pt, arc=2pt]
% Prompt e.g. The desk contains the necessary ingredients and a plate. Please observe the image and describe the items on the desk with very simple sentences.
% \end{tcolorbox}
% Next, based on the task requirements, we conduct feasibility analysis and task decomposition for ambiguous instructions. If the current scene cannot fulfill the instruction, we input the specific prompt into Qwen-VL 2.5 to generate feedback and suggestions.
% \begin{tcolorbox}[colback=gray!5!white, colframe=black, boxrule=0.5pt, arc=2pt]
% Prompt e.g. I want to eat spicy river seafood. Please recommend the ingredients that are still missing from the table and needed with very simple sentences.
% \end{tcolorbox}

Beyond understanding the current scene, CoT module incorporates future prediction through a generic frame sampling strategy.  At each time step t, We set a fixed time interval \( \Delta t = 30 \) frames and perform object detection using Qwen2.5-VL on the future frame at \( t' = \left\lfloor \frac{t}{\Delta t} \right\rfloor \times \Delta t + \Delta t \). Additionally, the robot’s state at frame \( t'' = t + \Delta t \) is converted into a textual description and appended to the CoT.

\subsection{Overall Framework of GraphCoT-VLA}

We adopt $\pi_0$ as the baseline. At each time step $t$, the robot receives multi-view images from the head, left and right wrist cameras: $I_{\text{head}, t}$, $I_{\text{left}, t}$, and $I_{\text{right}, t}$. Additionally, proprioception \(q_t\), an ambiguous instruction \(\mathcal{L}\), and a Pose-Object Graph \(\mathcal{G}_t\) are fed into model. Our model learns the conditional distribution of future actions as:
\begin{equation}
P(a_{t+1:t+\Delta t} \mid I_{\text{head}, t}, I_{\text{left}, t}, I_{\text{right}, t}, \mathcal{G}_t, q_t, \mathcal{L}).
\end{equation}

To model this distribution, we first encode the visual inputs \(\{I_{\text{head}, t}, I_{\text{left}, t}, I_{\text{right}, t}\}\) by vision transformer~\cite{dosovitskiy2020image}. The Pose-Object Graph \(\mathcal{G}_t\), representing the robot's configuration, is encoded using a graph encoder that treats each node as a graph token. Then the embeddings are concatenated with the tokenized language instruction \(\mathcal{L}\) to construct the input sequence for the model.

The input sequence is fed into a VLM built upon PaliGemma~\cite{beyer2024paligemma}. The output tokens from the VLM are divided into two parts: the first is used to autoregressively generate a CoT explanation, while the second is passed to an action expert module based on flow matching. This module also takes as input the robot state \(q_t\) and action noise \(\epsilon \sim \mathcal{N}(0, \sigma^2)\). The action expert then generates the predicted sequence of future actions \(\hat{A}_t = \hat{a}_{t+1:t+\Delta t}\).

\subsection{Co-Training with Dropout}

% 训练目标总共分成两部分，对生成的CoT的优化，以及对Action预测的优化。
% CoT优化:交叉熵损失被用来优化CoT的输出
% Action优化：Flow match loss

% 同时由于CoT的reasoning会大幅减慢推理速度，为了实现实时的action生成，我们采用了联合训练的方式，让模型具备推理的能力以及并实现实时输出。在训练时，在样本中以概率p选取样本对该样本中的cot输出进行dropout。然后将所有的样本用于训练，在测试时，则直接对cot部分作dropout，直接输出action结果，以实现高速的推理。

\paragraph{CoT Optimization.} 
To supervise the generation of CoT reasoning, we apply a standard cross-entropy loss~\cite{mao2023cross}:
\begin{equation}
\mathcal{L}_{\text{CoT}} = - \sum_{i=1}^{T_{\text{CoT}}} \log P(y_i \mid y_{<i}, o_t^{\text{CoT}}),
\end{equation}
where \( y_i \) denotes the \( i \)-th token in the ground-truth CoT sequence. $T_{\text{CoT}}$ denotes the number of tokens in the output CoT sequence, and $o_t^{CoT} = \{I_{\text{head}, t}, I_{\text{left}, t}, I_{\text{right}, t}, \mathcal{G}_t, q_t, \mathcal{L}\}$.

\paragraph{Action Optimization.} A conditional flow matching loss~\cite{lipman2022flow} is adopted  to learn a time-dependent denoising vector field. Given a future action sequence \( A_t = [a_{t+1}, \dots, a_{t+\Delta t}] \), we sample a time coefficient \( \tau \sim \text{Beta}(\alpha, \beta) \) and noise \( \epsilon \sim \mathcal{N}(0, I) \), and compute a perturbed action \( A_t^\tau \) as a weighted combination of the original action \( A_t \) and noise \( \epsilon \), given by \( A_t^\tau = \tau A_t + (1 - \tau) \epsilon \). The network is trained to predict the denoising vector \( \mathbf{u}(A_t^\tau \mid A_t) = \epsilon - A_t \), and loss defined as:
\begin{equation}
\mathcal{L}_{\text{action}} = \mathbb{E}_{A_t, \tau, \epsilon} \left[ \left\| \mathbf{v}_\theta(A_t^\tau, o_t^{\text{action}}) - (\epsilon - A_t) \right\|^2 \right],
\end{equation}
where \( \mathbf{v}_\theta \) is the learned vector field conditioned on the full observation. $o_t^{\text{action}}$ refers to the portion of the VLM output tokens following the CoT reasoning.

% To enable real-time inference, which is critical in robotic systems, we integrate a joint training strategy that balances reasoning capability and inference speed. During training, each sample undergoes CoT supervision dropout with probability \( p \), encouraging the model to learn to perform both with and without explicit reasoning. For samples without CoT dropout, the training loss is defined as a weighted sum of the CoT generation loss and the action prediction loss:
% \begin{equation}
% \mathcal{L}_{\text{total}} = \lambda_{\text{CoT}} \cdot \mathcal{L}_{\text{CoT}} + \lambda_{\text{action}} \cdot \mathcal{L}_{\text{action}}
% \end{equation}
% where \( \lambda_{\text{CoT}} \) and \( \lambda_{\text{action}} \) are hyperparameters balancing objectives. For samples with CoT dropout, the model is optimized using only the action prediction loss with \( \mathcal{L}_{\text{total}} = \mathcal{L}_{\text{action}} \).

To support real-time inference, we adopt a joint training strategy with CoT supervision dropout. Each sample is randomly dropped with probability \( p \), enabling the model to learn from both reasoning-guided and direct action prediction modes. The training loss is defined as:
\begin{equation}
\mathcal{L}_{\text{total}} = (1 - d) \cdot (\lambda_{\text{CoT}} \cdot \mathcal{L}_{\text{CoT}} + \lambda_{\text{action}} \cdot \mathcal{L}_{\text{action}}) + d \cdot \mathcal{L}_{\text{action}},
\end{equation}
where \( d \sim \text{Bernoulli}(p) \) indicates whether CoT supervision is dropped for the sample (\( d = 1 \) means dropout), and \( \lambda_{\text{CoT}}, \lambda_{\text{action}} \) are balancing weights.

During inference, the model adopts a hybrid reasoning strategy: CoT is generated only for the first frame to provide feedback and guidance, while subsequent frames skip reasoning and predict actions for robot control.

% To enable real-time inference, which is critical in robotic systems, we propose a joint training strategy that balances reasoning capability and inference speed. During training, each sample is subjected to CoT supervision dropout with a probability $p$, encouraging the model to learn to perform both with and without explicit reasoning. For samples without CoT dropout, the training loss is defined as a weighted sum of the CoT generation loss and the action prediction loss:
% \begin{equation}
% \mathcal{L}_{\text{total}} = \lambda_{\text{CoT}} \cdot \mathcal{L}_{\text{CoT}} + \lambda_{\text{action}} \cdot \mathcal{L}_{\text{action}},
% \end{equation}
% where $\lambda_{\text{CoT}}$ and $\lambda_{\text{action}}$ are hyperparameters balancing the two objectives. For samples with CoT dropout, only the action prediction loss is applied:
% \begin{equation}
% \mathcal{L}_{\text{total}} = \mathcal{L}_{\text{action}}.
% \end{equation}

% This training paradigm enables the model to retain the ability to perform explicit reasoning when needed, while maintaining high inference efficiency by directly predicting actions when reasoning is skipped.

% At inference time, we generate the Chain-of-Thought (CoT) output for the first frame to provide textual feedback and reasoning suggestions. Starting from the second frame, the model bypasses the CoT branch and directly predicts the action sequence from the VLM outputs, enabling fast, efficient, and smooth action inference.

\begin{table}[!b]
\centering
\begin{tabular}{ccccc}
\toprule
\multicolumn{5}{c}{\textbf{Food Preparation}} \\
\midrule
\textbf{Subtask} & \textbf{Egg} & \textbf{Tomato} & \textbf{Fish} & \textbf{Pepper} \\
\midrule
Subtask  1 & \ding{51} & \ding{51} & \ding{51} & \ding{51} \\
Subtask 2 & \ding{51} & \ding{51} &           & \ding{51} \\
Subtask 3  & \ding{51} & \ding{51} &           &           \\
\midrule
\multicolumn{5}{c}{\textbf{Outfit Selection}} \\
\midrule
\textbf{Subtask} & \textbf{Sweater} & \textbf{T-shirt} & \textbf{Shorts} &  \\
\midrule
Subtask 1 & \ding{51} & \ding{51} & \ding{51} &  \\
Subtask 2  & \ding{51} & \ding{51} &           &  \\
Subtask 3  & \ding{51} &           &           &  \\
\bottomrule
\end{tabular}
\caption{Available objects per subtask.}
\label{tab:available_objects}
\end{table}

\begin{table*}[t]
\centering
% \small
\begin{tabular}{c cccc cccc c}
\toprule
\multirow{2}{*}{\makecell[c]{\textbf{Method}}} & \multicolumn{4}{c}{\textbf{Food Preparation}} & \multicolumn{4}{c}{\textbf{Outfit Selection}} & \multirow{2}{*}{\makecell[c]{\textbf{No Task}\\\textbf{Confusion}}} \\
\cmidrule(lr){2-5} \cmidrule(lr){6-9}
& fish \& pepper & pepper & nothing & Avg. & T-shirt \& shorts & T-shirt & nothing & Avg. & \\
\midrule
% ACT~\cite{zhao2023learning} & 45 & 65 & 65 & 58.33 & 50 & 45 & 0 & 32.67 & \ding{55} \\
% Diffusion Policy~\cite{chi2023diffusion} & 35 & 65 & 60 & 53.33 & 45 & 30 & 5 & 26.67 & \ding{55} \\
% Octo fine-tuned~\cite{team2024octo} & 60 & 60 & \textbf{80} & 66.67 & 55 & 50 & 45 & 50.00 & \ding{55} \\
% $\pi_0$ fine-tuned~\cite{black2024pi_0} & 55 & 65 & 50 & 56.67 & 50 & 55 & 50 & 51.67 & \ding{55} \\
ACT & 45 & 65 & 65 & 58.33 & 50 & 45 & 0 & 32.67 & \ding{55} \\
Diffusion Policy & 35 & 65 & 60 & 53.33 & 45 & 30 & 5 & 26.67 & \ding{55} \\
Octo fine-tuned & 60 & 60 & \textbf{80} & 66.67 & 55 & 50 & 45 & 50.00 & \ding{55} \\
$\pi_0$ fine-tuned & 55 & 65 & 50 & 56.67 & 50 & 55 & 50 & 51.67 & \ding{55} \\
\midrule
Ours & \textbf{75} & \textbf{80} & 75 & \textbf{76.67} & \textbf{65} & \textbf{70} & \textbf{75} & \textbf{70.00} & \ding{51} \\
\bottomrule
\end{tabular}
\caption{\textbf{Comparison of our method and baselines.} Success rates (\%) for two tasks, the last column indicates task confusion.}
\label{tab:comparison}
\end{table*}

\begin{table*}[t]
\centering
\begin{tabular}{c cccc cccc c}
\toprule
\multirow{2}{*}{\textbf{Method}} & \multicolumn{4}{c}{\textbf{Food Preparation}} & \multicolumn{4}{c}{\textbf{Outfit Selection}} \\
\cmidrule(lr){2-5} \cmidrule(lr){6-9}
& fish \& pepper & pepper & nothing & Avg. & T-shirt \& shorts & T-shirt & nothing & Avg. \\
\midrule
w/o CoT\&Graph ($\pi_0$) & 55 & 65 & 50 & 56.67 & 50 & 55 & 50 & 51.67 \\
w/o PoseGraph & 70 & 80 & 75 & 75.00 & 55 & 60 & 55 & 56.67 \\
w/o CoT & 70 & 80 & 55 & 68.33 & 55 & 60 & 60 & 58.33 \\
\midrule
Ours & \textbf{75} & \textbf{80} & \textbf{75} & \textbf{76.67} & \textbf{65} & \textbf{70} & \textbf{75} & \textbf{70.00} \\
\bottomrule
\end{tabular}
\caption{\textbf{Ablation study.} Success rates (\%) of our model and variants on subtasks from two tasks.}
\label{tab:ablation}
\end{table*}

\section{Experiments}

\subsection{Experimental Setup}

\paragraph{Datasets.} We collect teleoperated data on a bimanual robot with 7 degrees of freedom per arm(Fig.~\ref{fig:overview}(b)), including RGB-D images from the head, neck, and hand-mounted cameras. As data streams run at different rates, with cameras at 30 \,Hz and proportional-derivative (PD) control at 150 \,Hz, we align all modalities using the head camera’s timestamp, actions are defined in joint angle space.

\paragraph{Task Descriptions.} The tasks, Food Preparation and Outfit Selection, are designed to evaluate the robot’s ability to perform under ambiguous instructions. The evaluation focuses on: (1) understanding ambiguous commands; (2) bimanual collaboration to test policy robustness and coordination; and (3) scenario diversity across subtasks to assess whether the model can distinguish tasks relying solely on visual and contextual information. Table~\ref{tab:available_objects} summarizes the varying object availability across subtasks for both Food Preparation and Outfit Selection, representing different situations the robot encounters.

To encourage generalization, we vary object placement during data collection, items are randomly translated 10\,cm and rotated up to $30^\circ$, clothing arrangements vary per subtask, with predefined permutations or uniform sampling across 10 hanger segments, we collect 100 demonstrations per subtask, totaling 600 for training.

% 对比的sota介绍
\paragraph{Baselines.} We compare our method with four state-of-the-art baselines:
\textbf{ACT}\cite{zhao2023learning}, a transformer-based trained from scratch to map images to actions;
\textbf{Diffusion Policy}\cite{chi2023diffusion}, an imitation learning method generating actions via diffusion, trained from scratch per task;
\textbf{Octo}\cite{team2024octo}, a pretrained transformer on 800k Open X-Embodiment~\cite{o2024open} episodes, fine-tuned with released weights;
\textbf{$\bm{\pi}_0$}\cite{black2024pi_0}, a general-purpose VLA model using both image and language inputs, also fine-tuned from released weights.

% \paragraph{Baselines.} Our method is compared with four state-of-the-art (SOTA) baselines: \textbf{ACT}\cite{zhao2023learning}, \textbf{Diffusion Policy}\cite{chi2023diffusion}, \textbf{Octo}\cite{team2024octo}, and \textbf{$\bm{\pi}_0$}\cite{black2024pi_0}.

\begin{figure*}[t!]
    \centering
    \includegraphics[width=0.95\textwidth]{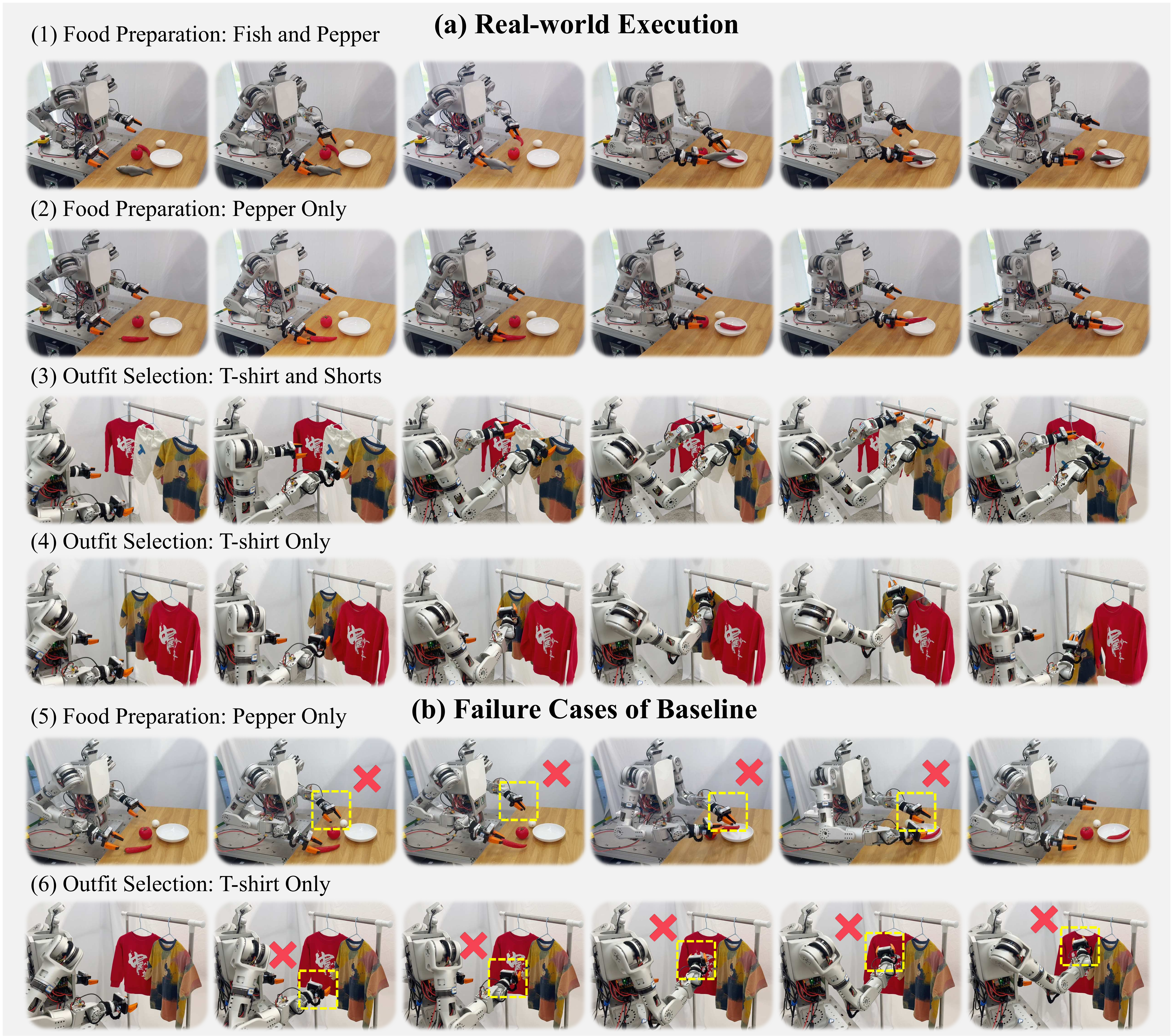}
    % \caption{\textbf{Snapshots of real-world executions and failure cases.} (a) Real-world Execution: Each row shows a task executed in the real world, with frames ordered chronologically from left to right. (b) Failure Cases of Baseline: The yellow boxes highlight the locations where the task failed.}
    \caption{\textbf{Real-world Executions.} Each row shows a task in order, with baseline failures highlighted in yellow.}
    \label{fig:vis_exe}
\end{figure*}

\begin{figure*}[t]
    \centering
    \includegraphics[width=0.95\textwidth]{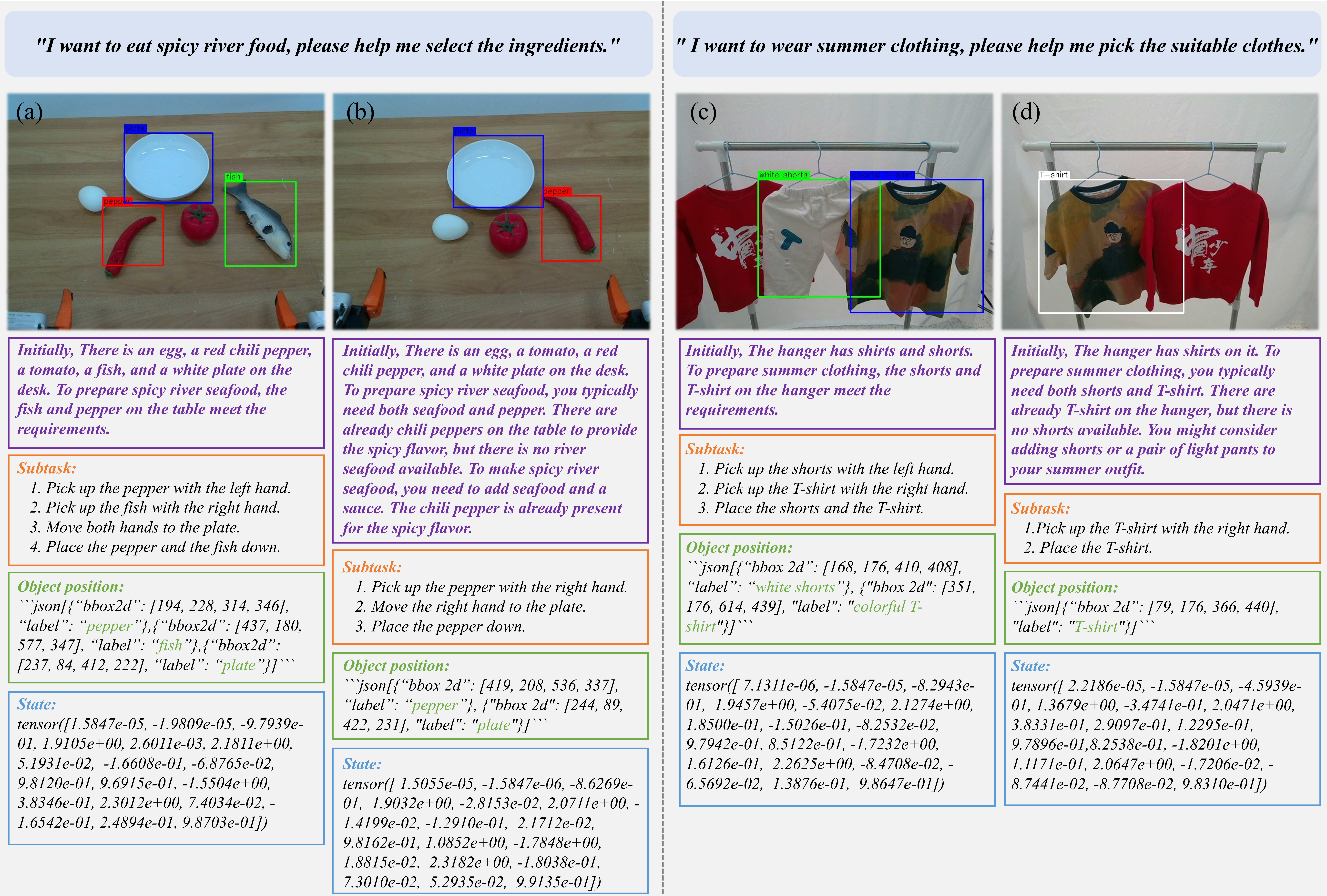}
    \caption{Visualization CoT reasoning from model inference.} 
    \label{fig:vis_cot}
\end{figure*}

\paragraph{Training and Evaluation Protocol.} 

We follow ACT's~\cite{zhao2023learning} task failure rate evaluation protocol, testing each task 20 times. A trial is successful only if the entire task is completed correctly. The details of the experimental setup and the criteria for identifying failure cases are provided in the appendix.

% The trained model is deployed for real-world reasoning and execution. All training and deployment are performed using 8 NVIDIA L40 GPUs. We follow ACT's~\cite{zhao2023learning} task failure rate evaluation protocol, testing each task 20 times. A trial is successful only if the entire task is completed correctly. Failure cases (Figure~\ref{fig:vis_exe}(b)) include: (1) Task confusion under ambiguous instructions, e.g., using one hand when both are required; (2) Grasp or placement failure with either hand, regardless of whether the task requires one or both; (3) Chaotic behavior~\cite{zhang20254d}: just like looking at a photo of a pendulum at its lowest point, it is hard to determine the direction of its swing. In our case, the robot continuously waves its hands without stopping.

\subsection{Comparative Results with SOTA Models}
Table~\ref{tab:comparison} compares our method with state-of-the-art approaches. In Food Preparation, Octo had the highest baseline accuracy, which we improved by 10\%. In Outfit Selection, $\pi_0$ led among baselines, and we outperformed it by 18.33\%. These results demonstrate that our approach offers higher accuracy and robustness, particularly in scenarios with task ambiguity. During real-world robot inference, as shown in Fig.~\ref{fig:vis_exe}, our method exhibits smoother body control. Notably, all baseline methods showed task confusion when handling ambiguous instructions. For example, as illustrated in Fig.~\ref{fig:vis_exe}(6), when the task required the robot to grasp a T-shirt, other methods incorrectly targeted a sweater. In contrast, our method correctly inferred the task intent and successfully executed the grasp, demonstrating superior task understanding and execution consistency.

\subsection{Ablation Studies}

\paragraph{3D Spatial Awareness through Pose-Object Graph.} As shown in Table ~\ref{tab:ablation}, the introduction of the Pose-Object Graph increases the success rate by up to 18.33\%. During the experiments, we observed that with graph reasoning (illustrated in Fig.~\ref{fig:vis_exe}(a)), the robot’s movements when grasping the target object become more decisive, coherent and natural. Furthermore, in the second task, which requires high end-effector precision to successfully grasp the hanger, the Pose-Object Graph leads to more accurate execution and greater operational stability. This demonstrates that the Pose-Object Graph improves the model’s accuracy and the fluency of action generation. 

% 结论：Pose-object Graph 的引入不仅提升了模型的准确率，还显著增强了动作生成的流畅性。

% 依据：如表格XXX所示，在 Food Preparation 任务中，引入 Pose-object Graph 显著提升了双手任务的成功率；在 Outfit Selection 任务中，单手和双手任务的成功率均提升了约 10%。在实验过程中我们观察到，在使用 Graph 推理的情况下（如图XXX(a)~所示），机器人在抓取目标物体时动作更加果断，整体行为更加连贯、自然。此外，在第二个任务中，夹爪需要具备较高的末端精度以成功抓取衣架，引入 Pose-object Graph 后，该任务的执行更加精准，展现出更高的操作稳定性。

\paragraph{Enhancing Comprehension and Reasoning with CoT.} As shown in Fig.~\ref{fig:vis_cot}(b), the CoT module enables the model to form a global understanding of the task based on varying environmental observations under ambiguous instructions. For instance, in the Food Selection task, when only chili peppers satisfy the given conditions, the model successfully provides reasonable suggestions. Additionally, it demonstrates preliminary sub-task planning abilities and the capacity to anticipate future steps. In addition, as shown in Table~\ref{tab:ablation}, the introduction of the CoT module leads to a significant improvement in the success rates. We also observed that the CoT module enhances the model’s ability to interpret and respond to chaotic or ambiguous situations. The baseline models often exhibit issues such as “insufficient or continuous waving”, which tends to make decisions based solely on current observations, lacking the capacity to reason about future task steps. With the CoT, the model demonstrates stronger task planning capabilities and a greater ability to predict future actions, resulting in more coherent behavior.

% 结论：引入 CoT（Chain-of-Thought）模块显著提升了模型的理解与推理能力。

% 依据：
% （1）如图 XXX 所示，在面对模糊指令的情况下，CoT 模块的输出能够根据不同的环境观测生成对任务的全局理解。例如，在“食物选择”任务中，当桌面上仅有辣椒满足条件时，模型成功地形成了对整体任务的理解，并给出了合理建议。同时，模型还具备初步的子任务规划能力，并展现出对未来步骤的预测能力。

% （2）此外，如表 XXX 所示，在两个任务的子任务中，引入 CoT 模块后，无论是单臂任务还是双臂任务，其成功率均有显著提升。我们还观察到，CoT 模块的引入增强了模型对混沌状态的理解与判断能力。相比于其他方法中常出现的“挥手次数不足”或“持续挥手不停”等问题，原始模型通常仅基于当前观测做出决策，缺乏对未来任务步骤的推理与想象，从而导致对任务时序的理解不充分。而加入 CoT 后，模型展现出更强的任务规划能力和对未来动作的预判能力，使行为表现更加合理和连贯。
\begin{figure}[!b]
  \centering
  \includegraphics[width=\linewidth]{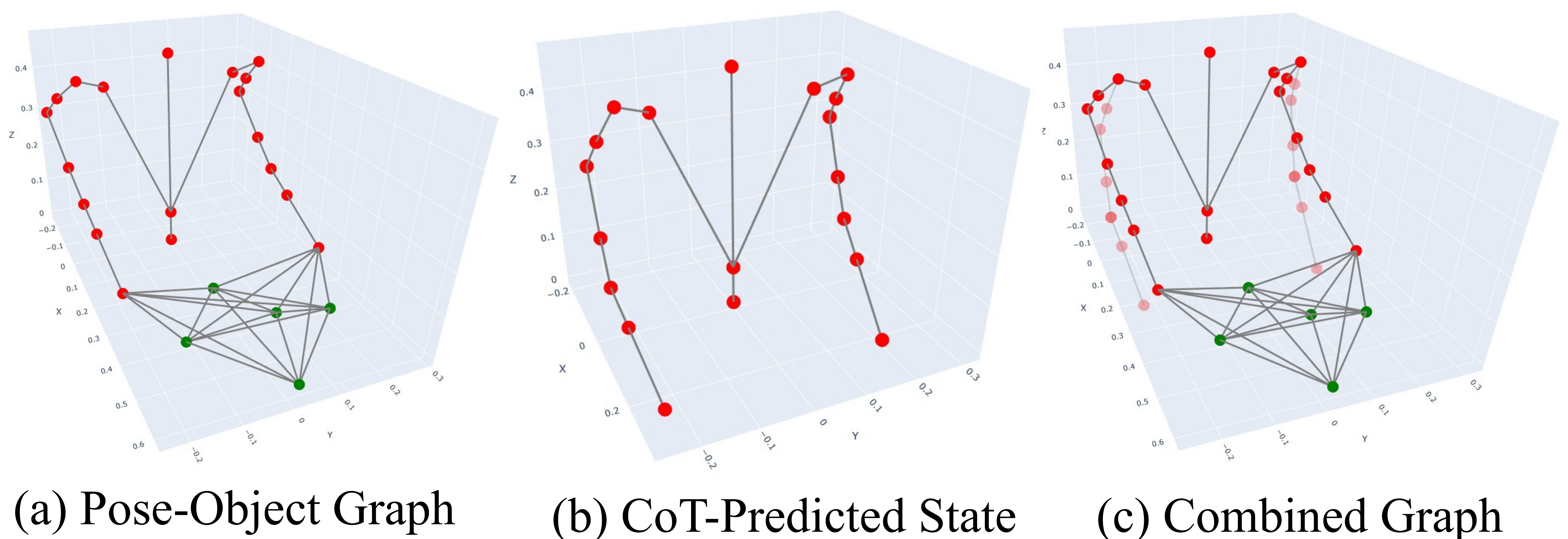}
  \caption{\textbf{Graph visualization.} (a) Pose-Object Graph: Red and green nodes represent robot joints and object points, respectively. (b) CoT Prediction: Robot state predction by CoT. (c) Combined Graph: Overlay of (a) and (b), with transparent nodes indicating CoT predictions.}
  \label{fig:vis_pose_graph}
\end{figure}

\subsection{Qualitative Visualization}

\paragraph{Pose-Object Graph Visualization.} We visualize its structure in Fig.~\ref{fig:vis_pose_graph}(a), where nodes represent robot joints or object centers in the world frame. Joints are connected via the kinematic chain, and the end-effector links to all objects to form a unified graph. This visualization clearly illustrates the spatial configuration between the robot and surrounding objects, demonstrating that the graph effectively encodes scene layout for downstream action generation.

\paragraph{Chain-of-thoughts visualization.} Fig.~\ref{fig:vis_cot} shows CoT’s reasoning under varying task conditions. In ambiguous scenes, where both fish and chili peppers are valid(Fig.~\ref{fig:vis_cot}(a), CoT captures global context (purple box) and generates appropriate subtasks (orange box). When only one valid option remains (Fig.~\ref{fig:vis_cot}(b)), it adapts its plan accordingly. The comparison between Fig.~\ref{fig:vis_cot}(c) and (d) further demonstrates that CoT adapts its interpretation and reasoning based on observed changes. These examples highlight CoT’s ability to interpret scene semantics and adjust decisions based on observation.

In addition to high-level reasoning, CoT accurately predicts low-level future states, including target object positions and the robot’s configuration. As shown in Fig.~\ref{fig:vis_cot}, the boxed areas mark predicted object locations that closely match the ground truth. Fig.~\ref{fig:vis_pose_graph}(c) further compares CoT’s predicted body state with actual motion, showing strong alignment in end-effector position and overall kinematics. These results confirm CoT’s capability in temporal reasoning and fine-grained state prediction.

% 图 XXX 展示了我们在机器人执行任务前的 CoT 推理结果。CoT 模块能够根据不同的环境观测，给出准确的全局任务理解反馈与子任务规划，从而显著提升模型的理解与推理能力。在模糊指令的条件下，图 XXX(a) 中的场景包含鱼和辣椒两种满足条件的物品，CoT 正确识别了这一环境状态，并生成了合理的全局理解（紫色框）及相应的子任务规划（橙色框）。相对地，在图 XXX(b) 中，当前观测中仅有辣椒满足条件，CoT 同样准确地识别出仅剩的目标，并给出针对性的反馈与建议，同时动态调整了子任务规划。类似地，在图 XXX(c) 中，满足模糊指令的目标为短裤与 T-shirt，而在图 XXX(d) 中仅有 T-shirt 满足条件，CoT 根据当前观测分别生成了正确的场景理解与相应的子任务并给出反馈，表现出良好的动态规划能力。

% 除了全局场景理解与子任务推理，CoT 还具备准确的低层级未来想象能力，包括目标物体位置的预测以及机器人自身状态的预测。为此，我们对其在目标位置与机器人状态上的预测结果进行了可视化分析。如图 XXX(a) 所示，图中框选区域为对目标物体位置的预测结果，能够看到目标不仅被准确识别，其空间位置也基本一致。图 XXX(b) 展示了 CoT 对机器人本体状态的预测情况。为了验证预测的准确性，我们将其与实时动作进行对比融合，结果如图 XXX(c) 所示，透明部分为 CoT 预测的状态，可以观察到机器人手臂的末端位置与真实位置接近，运动学形态也保持合理，验证了 CoT 在低层级时序理解推理有效性。

\subsection{Efficiency Study}

The model’s inference speed directly affects the smoothness of real-world robot execution. We measured the inference frequency of $\pi_0$ ten times, averaging about 10\,Hz (Table~\ref{tab:inference_frequency}). With CoT co-training on the same hardware, the first frame took slightly longer, but subsequent inference also ran at 10\,Hz, showing no added computational overhead.

% 模型的推理速度会实际机器人的执行丝滑程度，在实际真机测试过程中，我们测试了 $\pi_0$的推理频率，我们在两个任务上都进行了记录10次推理频率并且取平均，如表格xxx所示，$\pi_0$的推理频率大概在10HZ左右。同时在相同的硬件环境下，我们使用了CoT co-training的方法，除了第一帧需要更长时间推理，后面模型的推理速度约在10HZ，和$\pi_0$的推理速度保持一致，实验结果表明我们的方法在不降低模型推理频率。

\begin{table}[t]
\centering
\begin{tabular}{lc}
\toprule
\textbf{Model} & \textbf{Frequency} \\
\midrule
 $\pi_0$ &  $\approx$10 Hz \\
 Ours (co-training) &  $\approx$10 Hz \\
\bottomrule
\end{tabular}
\caption{Comparison of inference frequencies.}
\label{tab:inference_frequency}
\end{table}

\section{Conclusion}
In this work, we propose GraphCoT-VLA to address robotic manipulation tasks under ambiguous instructions. Specifically, we design a Pose-Object graph and a hierarchical CoT structure that enhance the model’s understanding and reasoning capabilities. Currently, our approach relies on future imagination and lacks memory of past temporal information, future work would explore incorporating historical sequences to enable more coherent long-term reasoning. Additionally, the current system is limited to static scenarios, we plan to extend it to mobile manipulation tasks, allowing robots to move and interact within dynamic environments.

\bibliography{aaai2026}

\begin{thebibliography}{30}
\providecommand{\natexlab}[1]{#1}

\bibitem[{Bai et~al.(2025)Bai, Chen, Liu, Wang, Ge, Song, Dang, Wang, Wang, Tang et~al.}]{bai2025qwen2}
Bai, S.; Chen, K.; Liu, X.; Wang, J.; Ge, W.; Song, S.; Dang, K.; Wang, P.; Wang, S.; Tang, J.; et~al. 2025.
\newblock Qwen2. 5-vl technical report.
\newblock \emph{arXiv preprint arXiv:2502.13923}.

\bibitem[{Beyer et~al.(2024)Beyer, Steiner, Pinto, Kolesnikov, Wang, Salz, Neumann, Alabdulmohsin, Tschannen, Bugliarello et~al.}]{beyer2024paligemma}
Beyer, L.; Steiner, A.; Pinto, A.~S.; Kolesnikov, A.; Wang, X.; Salz, D.; Neumann, M.; Alabdulmohsin, I.; Tschannen, M.; Bugliarello, E.; et~al. 2024.
\newblock Paligemma: A versatile 3b vlm for transfer.
\newblock \emph{arXiv preprint arXiv:2407.07726}.

\bibitem[{Black et~al.(2024)Black, Brown, Driess, Esmail, Equi, Finn, Fusai, Groom, Hausman, Ichter et~al.}]{black2024pi_0}
Black, K.; Brown, N.; Driess, D.; Esmail, A.; Equi, M.; Finn, C.; Fusai, N.; Groom, L.; Hausman, K.; Ichter, B.; et~al. 2024.
\newblock $pi\_0$: A Vision-Language-Action Flow Model for General Robot Control.
\newblock \emph{arXiv preprint arXiv:2410.24164}.

\bibitem[{Brohan et~al.(2022)Brohan, Brown, Carbajal, Chebotar, Dabis, Finn, Gopalakrishnan, Hausman, Herzog, Hsu et~al.}]{brohan2022rt}
Brohan, A.; Brown, N.; Carbajal, J.; Chebotar, Y.; Dabis, J.; Finn, C.; Gopalakrishnan, K.; Hausman, K.; Herzog, A.; Hsu, J.; et~al. 2022.
\newblock Rt-1: Robotics transformer for real-world control at scale.
\newblock \emph{arXiv preprint arXiv:2212.06817}.

\bibitem[{Cheang et~al.(2024)Cheang, Chen, Jing, Kong, Li, Li, Liu, Wu, Xu, Yang et~al.}]{cheang2024gr}
Cheang, C.-L.; Chen, G.; Jing, Y.; Kong, T.; Li, H.; Li, Y.; Liu, Y.; Wu, H.; Xu, J.; Yang, Y.; et~al. 2024.
\newblock Gr-2: A generative video-language-action model with web-scale knowledge for robot manipulation.
\newblock \emph{arXiv preprint arXiv:2410.06158}.

\bibitem[{Chen et~al.(2025)Chen, Belkhale, Mirchandani, Mees, Driess, Pertsch, and Levine}]{chen2025training}
Chen, W.; Belkhale, S.; Mirchandani, S.; Mees, O.; Driess, D.; Pertsch, K.; and Levine, S. 2025.
\newblock Training Strategies for Efficient Embodied Reasoning.
\newblock \emph{arXiv preprint arXiv:2505.08243}.

\bibitem[{Cheng et~al.(2024)Cheng, Song, Ge, Liu, Wang, and Shan}]{cheng2024yolo}
Cheng, T.; Song, L.; Ge, Y.; Liu, W.; Wang, X.; and Shan, Y. 2024.
\newblock Yolo-world: Real-time open-vocabulary object detection.
\newblock In \emph{Proceedings of the IEEE/CVF conference on computer vision and pattern recognition}, 16901--16911.

\bibitem[{Chi et~al.(2023)Chi, Xu, Feng, Cousineau, Du, Burchfiel, Tedrake, and Song}]{chi2023diffusion}
Chi, C.; Xu, Z.; Feng, S.; Cousineau, E.; Du, Y.; Burchfiel, B.; Tedrake, R.; and Song, S. 2023.
\newblock Diffusion policy: Visuomotor policy learning via action diffusion.
\newblock \emph{The International Journal of Robotics Research}, 02783649241273668.

\bibitem[{Din et~al.(2025)Din, Akram, Saoud, Rosell, and Hussain}]{din2025vision}
Din, M.~U.; Akram, W.; Saoud, L.~S.; Rosell, J.; and Hussain, I. 2025.
\newblock Vision Language Action Models in Robotic Manipulation: A Systematic Review.
\newblock \emph{arXiv preprint arXiv:2507.10672}.

\bibitem[{Dosovitskiy et~al.(2020)Dosovitskiy, Beyer, Kolesnikov, Weissenborn, Zhai, Unterthiner, Dehghani, Minderer, Heigold, Gelly et~al.}]{dosovitskiy2020image}
Dosovitskiy, A.; Beyer, L.; Kolesnikov, A.; Weissenborn, D.; Zhai, X.; Unterthiner, T.; Dehghani, M.; Minderer, M.; Heigold, G.; Gelly, S.; et~al. 2020.
\newblock An image is worth 16x16 words: Transformers for image recognition at scale.
\newblock \emph{arXiv preprint arXiv:2010.11929}.

\bibitem[{Kim et~al.(2024)Kim, Pertsch, Karamcheti, Xiao, Balakrishna, Nair, Rafailov, Foster, Lam, Sanketi et~al.}]{kim2024openvla}
Kim, M.~J.; Pertsch, K.; Karamcheti, S.; Xiao, T.; Balakrishna, A.; Nair, S.; Rafailov, R.; Foster, E.; Lam, G.; Sanketi, P.; et~al. 2024.
\newblock Openvla: An open-source vision-language-action model.
\newblock \emph{arXiv preprint arXiv:2406.09246}.

\bibitem[{Lipman et~al.(2022)Lipman, Chen, Ben-Hamu, Nickel, and Le}]{lipman2022flow}
Lipman, Y.; Chen, R.~T.; Ben-Hamu, H.; Nickel, M.; and Le, M. 2022.
\newblock Flow matching for generative modeling.
\newblock \emph{arXiv preprint arXiv:2210.02747}.

\bibitem[{Liu et~al.(2024)Liu, Wu, Li, Tan, Chen, Wang, Xu, Su, and Zhu}]{liu2024rdt}
Liu, S.; Wu, L.; Li, B.; Tan, H.; Chen, H.; Wang, Z.; Xu, K.; Su, H.; and Zhu, J. 2024.
\newblock Rdt-1b: a diffusion foundation model for bimanual manipulation.
\newblock \emph{arXiv preprint arXiv:2410.07864}.

\bibitem[{Mao, Mohri, and Zhong(2023)}]{mao2023cross}
Mao, A.; Mohri, M.; and Zhong, Y. 2023.
\newblock Cross-entropy loss functions: Theoretical analysis and applications.
\newblock In \emph{International conference on Machine learning}, 23803--23828. pmlr.

\bibitem[{O’Neill et~al.(2024)O’Neill, Rehman, Maddukuri, Gupta, Padalkar, Lee, Pooley, Gupta, Mandlekar, Jain et~al.}]{o2024open}
O’Neill, A.; Rehman, A.; Maddukuri, A.; Gupta, A.; Padalkar, A.; Lee, A.; Pooley, A.; Gupta, A.; Mandlekar, A.; Jain, A.; et~al. 2024.
\newblock Open x-embodiment: Robotic learning datasets and rt-x models: Open x-embodiment collaboration 0.
\newblock In \emph{2024 IEEE International Conference on Robotics and Automation (ICRA)}, 6892--6903. IEEE.

\bibitem[{Qu et~al.(2025)Qu, Song, Chen, Yao, Ye, Ding, Wang, Gu, Zhao, Wang et~al.}]{qu2025spatialvla}
Qu, D.; Song, H.; Chen, Q.; Yao, Y.; Ye, X.; Ding, Y.; Wang, Z.; Gu, J.; Zhao, B.; Wang, D.; et~al. 2025.
\newblock Spatialvla: Exploring spatial representations for visual-language-action model.
\newblock \emph{arXiv preprint arXiv:2501.15830}.

\bibitem[{Scarselli et~al.(2008)Scarselli, Gori, Tsoi, Hagenbuchner, and Monfardini}]{scarselli2008graph}
Scarselli, F.; Gori, M.; Tsoi, A.~C.; Hagenbuchner, M.; and Monfardini, G. 2008.
\newblock The graph neural network model.
\newblock \emph{IEEE transactions on neural networks}, 20(1): 61--80.

\bibitem[{Team et~al.(2024)Team, Ghosh, Walke, Pertsch, Black, Mees, Dasari, Hejna, Kreiman, Xu et~al.}]{team2024octo}
Team, O.~M.; Ghosh, D.; Walke, H.; Pertsch, K.; Black, K.; Mees, O.; Dasari, S.; Hejna, J.; Kreiman, T.; Xu, C.; et~al. 2024.
\newblock Octo: An open-source generalist robot policy.
\newblock \emph{arXiv preprint arXiv:2405.12213}.

\bibitem[{Vaswani et~al.(2017)Vaswani, Shazeer, Parmar, Uszkoreit, Jones, Gomez, Kaiser, and Polosukhin}]{vaswani2017attention}
Vaswani, A.; Shazeer, N.; Parmar, N.; Uszkoreit, J.; Jones, L.; Gomez, A.~N.; Kaiser, {\L}.; and Polosukhin, I. 2017.
\newblock Attention is all you need.
\newblock \emph{Advances in neural information processing systems}, 30.

\bibitem[{Wang(2025)}]{wang2025roboflamingo}
Wang, S. 2025.
\newblock Roboflamingo-plus: Fusion of depth and rgb perception with vision-language models for enhanced robotic manipulation.
\newblock \emph{arXiv preprint arXiv:2503.19510}.

\bibitem[{Wei et~al.(2022)Wei, Wang, Schuurmans, Bosma, Xia, Chi, Le, Zhou et~al.}]{wei2022chain}
Wei, J.; Wang, X.; Schuurmans, D.; Bosma, M.; Xia, F.; Chi, E.; Le, Q.~V.; Zhou, D.; et~al. 2022.
\newblock Chain-of-thought prompting elicits reasoning in large language models.
\newblock \emph{Advances in neural information processing systems}, 35: 24824--24837.

\bibitem[{Wen et~al.(2024)Wen, Zhu, Zhu, Tang, Li, Zhou, Li, Liu, Peng, Shen et~al.}]{wen2024diffusion}
Wen, J.; Zhu, M.; Zhu, Y.; Tang, Z.; Li, J.; Zhou, Z.; Li, C.; Liu, X.; Peng, Y.; Shen, C.; et~al. 2024.
\newblock Diffusion-VLA: Generalizable and Interpretable Robot Foundation Model via Self-Generated Reasoning.
\newblock \emph{arXiv preprint arXiv:2412.03293}.

\bibitem[{Wen et~al.(2025{\natexlab{a}})Wen, Zhu, Li, Tang, Shen, and Feng}]{wen2025dexvla}
Wen, J.; Zhu, Y.; Li, J.; Tang, Z.; Shen, C.; and Feng, F. 2025{\natexlab{a}}.
\newblock Dexvla: Vision-language model with plug-in diffusion expert for general robot control.
\newblock \emph{arXiv preprint arXiv:2502.05855}.

\bibitem[{Wen et~al.(2025{\natexlab{b}})Wen, Zhu, Li, Zhu, Tang, Wu, Xu, Liu, Cheng, Shen et~al.}]{wen2025tinyvla}
Wen, J.; Zhu, Y.; Li, J.; Zhu, M.; Tang, Z.; Wu, K.; Xu, Z.; Liu, N.; Cheng, R.; Shen, C.; et~al. 2025{\natexlab{b}}.
\newblock Tinyvla: Towards fast, data-efficient vision-language-action models for robotic manipulation.
\newblock \emph{IEEE Robotics and Automation Letters}.

\bibitem[{Wu et~al.(2023)Wu, Jing, Cheang, Chen, Xu, Li, Liu, Li, and Kong}]{wu2023unleashing}
Wu, H.; Jing, Y.; Cheang, C.; Chen, G.; Xu, J.; Li, X.; Liu, M.; Li, H.; and Kong, T. 2023.
\newblock Unleashing large-scale video generative pre-training for visual robot manipulation.
\newblock \emph{arXiv preprint arXiv:2312.13139}.

\bibitem[{Zawalski et~al.(2024)Zawalski, Chen, Pertsch, Mees, Finn, and Levine}]{zawalski2024robotic}
Zawalski, M.; Chen, W.; Pertsch, K.; Mees, O.; Finn, C.; and Levine, S. 2024.
\newblock Robotic control via embodied chain-of-thought reasoning.
\newblock \emph{arXiv preprint arXiv:2407.08693}.

\bibitem[{Zhang and Yan(2023)}]{zhang2023crossformer}
Zhang, Y.; and Yan, J. 2023.
\newblock Crossformer: Transformer utilizing cross-dimension dependency for multivariate time series forecasting.
\newblock In \emph{The eleventh international conference on learning representations}.

\bibitem[{Zhao et~al.(2025)Zhao, Lu, Kim, Fu, Zhang, Wu, Li, Ma, Han, Finn et~al.}]{zhao2025cot}
Zhao, Q.; Lu, Y.; Kim, M.~J.; Fu, Z.; Zhang, Z.; Wu, Y.; Li, Z.; Ma, Q.; Han, S.; Finn, C.; et~al. 2025.
\newblock Cot-vla: Visual chain-of-thought reasoning for vision-language-action models.
\newblock In \emph{Proceedings of the Computer Vision and Pattern Recognition Conference}, 1702--1713.

\bibitem[{Zhao et~al.(2023)Zhao, Kumar, Levine, and Finn}]{zhao2023learning}
Zhao, T.~Z.; Kumar, V.; Levine, S.; and Finn, C. 2023.
\newblock Learning fine-grained bimanual manipulation with low-cost hardware.
\newblock \emph{arXiv preprint arXiv:2304.13705}.

\bibitem[{Zitkovich et~al.(2023)Zitkovich, Yu, Xu, Xu, Xiao, Xia, Wu, Wohlhart, Welker, Wahid et~al.}]{zitkovich2023rt}
Zitkovich, B.; Yu, T.; Xu, S.; Xu, P.; Xiao, T.; Xia, F.; Wu, J.; Wohlhart, P.; Welker, S.; Wahid, A.; et~al. 2023.
\newblock Rt-2: Vision-language-action models transfer web knowledge to robotic control.
\newblock In \emph{Conference on Robot Learning}, 2165--2183. PMLR.

\end{thebibliography}
\end{document}